\documentclass[runningheads]{llncs}

\usepackage[T1]{fontenc}
\usepackage{amsmath}
\usepackage{amssymb}
\usepackage{microtype}
\usepackage{graphicx}
\usepackage{adjustbox}
\usepackage{booktabs}
\usepackage{array}
\usepackage{float}
\usepackage{flafter}
\usepackage[hidelinks]{hyperref}
\begin{document}

\title{TASTE: Time-Aware Sequential Evidence Selection with Adaptive Stopping for Time-Sensitive Question Answering}
\titlerunning{TASTE: Time-Aware Sequential Evidence Selection}

\author{Lulu Chen\inst{1}$^{\dagger}$ \and
Yiliu Yang\inst{2}$^{\dagger}$ \and
Yue Sun\inst{1} \and
Siyu Wang\inst{1}$^{\star}$}
\authorrunning{L. Chen et al.}
\institute{Gusu Laboratory of Materials, Suzhou, China\\
\email{\{chenlulu2021,sunyue2021,wangsiyu2022\}@gusulab.ac.cn}
\and
Jiangnan University, Wuxi, China\\
\email{yyangyiliu@outlook.com}}

\maketitle
{\makeatletter
\renewcommand\@makefntext[1]{\par\noindent #1}%
\makeatother
\footnotetext{$^{\dagger}$Equal contribution.}%
\footnotetext{$^{\star}$Corresponding author.}}

\begin{abstract}
Time-sensitive question answering needs evidence that supports a complete answer under temporal constraints. Temporally relevant candidates may repeat the same information while missing other answer parts, and fixed top-$k$ selection cannot adapt to evidence sufficiency. We propose TASTE (Time-Aware Sequential Evidence Selection), which jointly learns what evidence to select next and when to stop. Its policy uses semantic and temporal features together with the selected evidence to favor complementary support, stop when support is sufficient, or return an empty set when no supporting evidence is available. Supervised warm-start and reinforcement learning encourage sufficient, temporally matched, compact evidence sets. Experiments on two fixed-pool datasets show higher sufficiency and useful rates than ranking baselines at matched evidence counts, lower temporal conflict rates, and cross-dataset transfer. TASTE returns empty sets for 80.9\% of unanswerable test questions in TempPool-TimeQA.

\end{abstract}

\section{Introduction}
\label{sec:intro}

Many real-world facts change over time, such as who holds a position.
Time-sensitive questions therefore require answers valid for the queried time~\cite{chen2021timeqa,ning2020torque,zhang2021situatedqa}.
Retrieval-augmented generation (RAG) supplies external evidence to address outdated or missing model knowledge~\cite{lewis2020rag,vu2024freshllms}. This makes it useful for time-sensitive QA.
However, evidence relevant to the question's topic need not be valid for the queried period. When a question spans multiple periods or requires multiple answers, a single relevant passage is often insufficient.
Time-sensitive RAG needs a compact evidence set that matches the topic and time constraint and supports a complete answer.

\begin{figure}[t]
\centering
\includegraphics[width=0.9\linewidth]{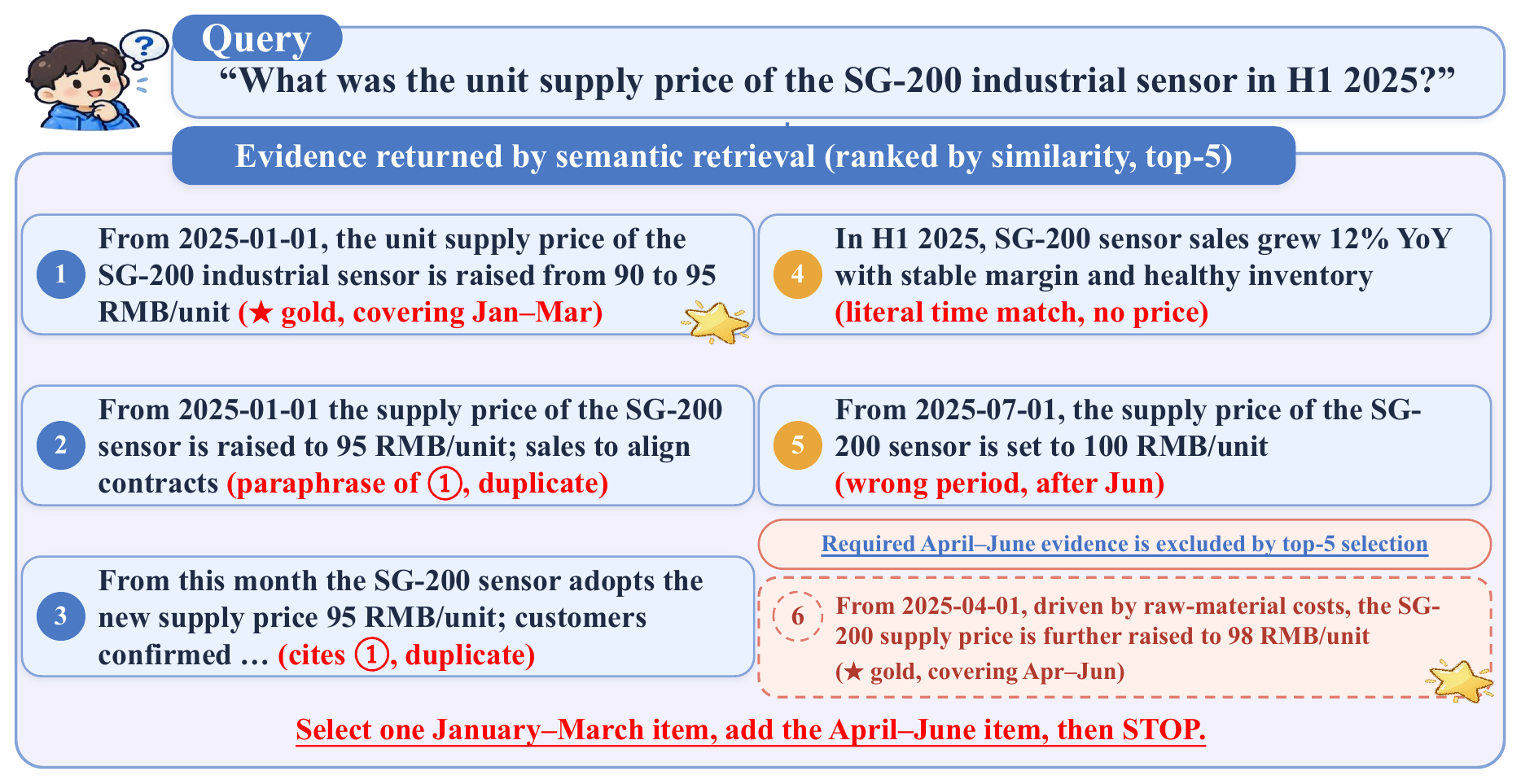}
\caption{Illustrative example of fixed top-$k$ selection for time-sensitive QA.}
\label{fig:problem}
\end{figure}

Many methods rank candidate passages and keep the top-$k$ for a preset $k$. Semantic ranking methods score each passage by its relevance to the question~\cite{chen2024bge,izacard2022contriever,zhang2025qwen3}. Time-aware methods such as MRAG also use temporal relevance~\cite{zhang2025mrag}. However, a fixed $k$ may keep redundant passages or miss necessary evidence. Set selection methods consider how passages jointly support the answer~\cite{lee2025setselection}. Adaptive-count methods adjust how many passages to return for each question~\cite{sun2025dynamicrag,taguchi2025adaptivek}. For time-sensitive QA, these decisions are closely linked. The value of another passage depends on its temporal validity and the evidence already selected. The selector must decide whether the passage adds needed support or whether selection should stop.

Figure~\ref{fig:problem} illustrates why time-sensitive QA must consider both which evidence to select and when to stop. The question asks for the SG-200 sensor's price in the first half of 2025. The top-five passages include repeated January evidence, an irrelevant summary, and a July update, but miss the April--June evidence ranked sixth. Increasing $k$ to six recovers this evidence but retains redundant, irrelevant, and temporally invalid passages. Two challenges follow. \textbf{(1) Selecting evidence that adds missing support.} After selecting January--March evidence, the selector should add April--June evidence. \textbf{(2) Stopping at the right time.} Here, the selector should stop after selecting one supporting item per period. If no supporting evidence exists, it should stop immediately and return an empty set.

We propose \textbf{TASTE} (Time-Aware Sequential Evidence Selection) to jointly learn \textbf{what to select next and when to stop under temporal constraints}. TASTE combines semantic and temporal features with attention across candidates. Its policy uses the selected evidence, temporal hits, redundancy, and remaining budget to choose a candidate or STOP. Supervised warm-start initializes candidate scoring. Reinforcement learning then jointly trains evidence selection and stopping. We construct two fixed-pool datasets from TimeQA~\cite{chen2021timeqa} and TempRAGEval's SituatedQA subset~\cite{zhang2025mrag}. At matched evidence counts, TASTE achieves higher sufficiency and useful rates than ranking baselines on both datasets. It also has lower temporal conflict rates than all baselines. TASTE returns empty sets for 80.9\% of unanswerable test questions in TempPool-TimeQA.

\section{Related Work}
\label{sec:related}

\paragraph{Time-sensitive QA datasets.}
TimeQA~\cite{chen2021timeqa}, SituatedQA~\cite{zhang2021situatedqa}, and TempRAGEval~\cite{zhang2025mrag} evaluate time-sensitive QA. Other resources study temporal knowledge~\cite{dhingra2022templama}, reasoning over text or knowledge graphs~\cite{wei2023menatqa,saxena2021tkqg}, and evolving information~\cite{jang2022temporalwiki,kasai2023realtimeqa,liska2022streamingqa}.

\paragraph{Time-aware retrieval and reranking.}
TempRetriever~\cite{abdallah2026tempretriever} embeds query dates and document timestamps. TempRALM~\cite{gade2025tempralm} combines semantic and temporal relevance. TA-RAG~\cite{lau2025tarag} parses temporal windows. MRAG~\cite{zhang2025mrag} combines semantic and temporal sentence scores. TimeR4~\cite{qian2024timer4} retrieves temporal knowledge-graph facts.

\paragraph{Evidence selection for RAG.}
DynamicRAG~\cite{sun2025dynamicrag} learns ordering and counts through reinforcement learning. Adaptive-$k$~\cite{taguchi2025adaptivek} determines counts from similarity scores. SetR~\cite{lee2025setselection} selects passages that jointly meet information needs. Reflective and multi-step methods control retrieval~\cite{asai2024selfrag,song2025r1searcher,trivedi2023ircot}. Sufficient Context~\cite{joren2025sufficient} studies context sufficiency and abstention. TASTE learns selection and STOP from temporal features and selected-set state within a fixed pool.

\section{Method}
\label{sec:method}

\subsection{Task Definition}

\paragraph{Input.}
The input is a time-sensitive question $q$ and a fixed pool of $N$ candidate passages, $\mathcal{C}=\{c_i\}_{i=1}^{N}$. The question is split into main content (MC) and a time constraint (TC). MC expresses the non-temporal content of the question. TC is a nonempty set of query intervals $\mathcal{T}_q=\{[l_k,r_k]\}_{k=1}^{|\mathcal{T}_q|}$. Finite endpoints use day granularity; left and right endpoints may be $-\infty$ and $+\infty$, respectively. Each candidate $c_i$ contains text and a set of extracted time intervals $\mathcal{T}_i=\{[l_{i,j},r_{i,j}]\}_{j=1}^{|\mathcal{T}_i|}$. We set $\mathcal{T}_i=\varnothing$ when no time interval can be extracted.

\paragraph{Output.}
The output is an evidence set $\mathcal{S}\subseteq\mathcal{C}$, with $|\mathcal{S}|\leq K$, where $K$ is the maximum number of selected items. The output may be empty.

\paragraph{Objective.}
The goal is to select a compact set that supports all required answer parts under TC, while limiting redundancy and temporal conflicts. If the pool contains no supporting evidence under TC, the selector should return an empty set.
\subsection{Overview}
\label{sec:overview}

\begin{figure}[t]
\centering
\includegraphics[width=0.9\textwidth]{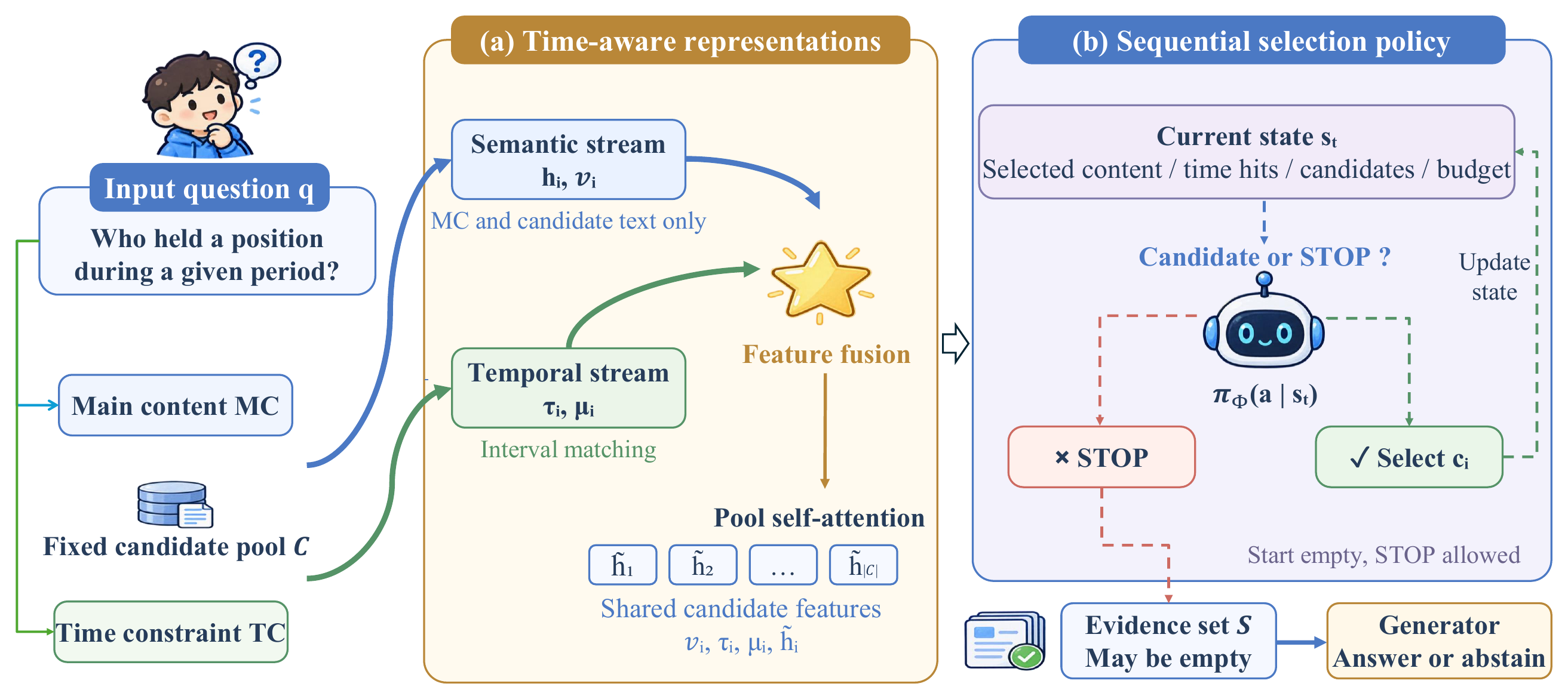}
\caption{TASTE: (a) time-aware representations; (b) sequential selection. The downstream generator is shown for context and is not evaluated.}
\label{fig:arch}
\end{figure}

Figure~\ref{fig:arch} shows two modules. The \textbf{time-aware representation module} combines text, semantic, and temporal features through pool-level self-attention. The \textbf{sequential decision module} uses these representations and the selected-set state to score candidates and STOP. Supervised warm-start trains candidate scoring; reinforcement learning then trains selection and stopping.

\subsection{Time-Aware Representation Module}

\paragraph{Semantic relevance.}
The encoder $\operatorname{Enc}$ maps text to an $\ell_2$-normalized vector in $\mathbb{R}^{D}$. We encode candidate $c_i$ and MC, then compute their cosine similarity:
\begin{equation}
\mathbf{h}_i=\operatorname{Enc}(c_i), \qquad
\mathbf{h}_{\mathrm{MC}}=\operatorname{Enc}(\mathrm{MC}), \qquad
v_i=\cos(\mathbf{h}_i,\mathbf{h}_{\mathrm{MC}}).
\label{eq:semantic_relevance}
\end{equation}
We encode MC rather than the full question to separate semantic and temporal matching. Candidate text retains its temporal expressions.

\paragraph{Temporal relevance.}
For candidate $c_i$ and query interval $[l_k,r_k]$, we define the temporal hit:
\begin{equation}
\tau_{i,k}=
\begin{cases}
0, & \text{if } \mathcal{T}_i=\varnothing,\\
1, & \text{if } \exists j\in\{1,\ldots,|\mathcal{T}_i|\}:
[l_{i,j},r_{i,j}]\cap[l_k,r_k]\neq\varnothing,\\
-1, & \text{otherwise}.
\end{cases}
\label{eq:temporal_hit}
\end{equation}
Here, $1$ denotes overlap, $-1$ denotes no overlap, and $0$ denotes missing time information. Overlap alone does not imply answer support or full interval coverage.

We collect the hits in $\boldsymbol{\tau}_i=[\tau_{i,1},\ldots,\tau_{i,|\mathcal{T}_q|}]^\top$. Their maximum is $\tau_i^{\max}=\max_k\tau_{i,k}$, where $1\leq k\leq|\mathcal{T}_q|$. This maximum records whether the candidate overlaps any query interval. We pad $\boldsymbol{\tau}_i$ to length $M$ for network input, giving $\bar{\boldsymbol{\tau}}_i\in\mathbb{R}^{M}$. Padding is excluded from the maximum.

The flag $\mu_i=\mathbb{I}[\mathcal{T}_i=\varnothing]$ explicitly identifies untimed candidates. Here, $\mathbb{I}[\cdot]$ is the indicator function.

\paragraph{Feature fusion.}
We concatenate the candidate vector, semantic score, temporal hits, and missing-time flag. A linear projection gives
\begin{equation}
\mathbf{x}_i=
\mathbf{W}
[\mathbf{h}_i;v_i;\bar{\boldsymbol{\tau}}_i;\mu_i],
\label{eq:feature_fusion}
\end{equation}
where $[\,;\,]$ denotes concatenation, $\mathbf{W}\in\mathbb{R}^{D\times(D+M+2)}$, and $\mathbf{x}_i\in\mathbb{R}^{D}$.

We apply multi-head self-attention across the pool~\cite{vaswani2017attention}. A residual connection and layer normalization give
\begin{equation}
\widetilde{\mathbf{h}}_i=
\operatorname{LN}
\bigl(\mathbf{x}_i+\operatorname{Attn}_i(\mathbf{X})\bigr),
\label{eq:pool_attention}
\end{equation}
where
$\mathbf{X}=
[\mathbf{x}_1^\top;\ldots;\mathbf{x}_N^\top]
\in\mathbb{R}^{N\times D}$.
Here, $\operatorname{Attn}_i(\mathbf{X})\in\mathbb{R}^{D}$ is the attention output for candidate $i$, written as a column vector. LN denotes layer normalization. Self-attention lets each candidate incorporate information from the pool.

These representations do not depend on the selected set. We compute them once per question and reuse them at every step.
\subsection{Sequential Decision Module}

\paragraph{Decision state.}
The state at step $t$ is $s_t=(\mathcal{S}_t,\mathcal{R}_t,\mathbf{g}_t,\mathbf{m}_t,b_t)$,
where $\mathcal{S}_t$ is the selected set,
$\mathcal{R}_t$ is the remaining set,
$\mathbf{g}_t$ represents the selected content,
$\mathbf{m}_t$ records temporal hits, and $b_t$ is the remaining budget.
Initially, $\mathcal{S}_0=\varnothing$.
The remaining candidates and budget are
$\mathcal{R}_t=\mathcal{C}\setminus\mathcal{S}_t$ and $b_t=K-|\mathcal{S}_t|$.

We represent the selected content by the mean of the selected encoder representations:
\begin{equation}
\mathbf{g}_t=
\begin{cases}
\mathbf{0}, & \text{if } \mathcal{S}_t=\varnothing,\\
\displaystyle
\frac{1}{|\mathcal{S}_t|}
\sum_{c_i\in\mathcal{S}_t}\mathbf{h}_i,
& \text{otherwise}.
\end{cases}
\label{eq:selected_content}
\end{equation}

For each query interval $k$, we record whether any selected candidate matches it:
\begin{equation}
m_{t,k}=
\begin{cases}
0, & \text{if } \mathcal{S}_t=\varnothing,\\
1, & \text{if } \exists c_i\in\mathcal{S}_t:
\tau_{i,k}=1,\\
-1, & \text{otherwise}.
\end{cases}
\label{eq:selected_temporal_hits}
\end{equation}
We collect these values into
$\mathbf{m}_t=
[m_{t,1},\ldots,m_{t,|\mathcal{T}_q|}]^\top$.
Here, $-1$ means no temporal hit, including when all selected candidates are untimed. We pad $\mathbf{m}_t$ to length $M$ as $\bar{\mathbf{m}}_t$ for network input.

\paragraph{Candidate scoring.}
Each remaining candidate is scored using its own features and its relation to the selected evidence. We measure redundancy by the maximum cosine similarity to a selected candidate:
\begin{equation}
d_{t,i}=
\begin{cases}
0, & \text{if } \mathcal{S}_t=\varnothing,\\
\displaystyle
\max_{c_{i'}\in\mathcal{S}_t}
\cos(\mathbf{h}_i,\mathbf{h}_{i'}),
& \text{otherwise}.
\end{cases}
\label{eq:redundancy}
\end{equation}

We also compute a potential complementarity feature:
\begin{equation}
u_{t,i}=
\begin{cases}
\max(v_i,0)\bigl(1-\max(d_{t,i},0)\bigr),
& \text{if } \tau_i^{\max}=1,\\
0, & \text{otherwise}.
\end{cases}
\label{eq:potential_complementarity}
\end{equation}
This heuristic favors relevant, temporally matched, non-redundant candidates without directly identifying missing answer parts. A zero value does not exclude the candidate; other features can still support its selection.

For each $c_i\in\mathcal{R}_t$, the scoring network receives
\begin{equation}
\boldsymbol{\phi}_{t,i}=
[v_i;\bar{\boldsymbol{\tau}}_i;\mu_i;
\mathbf{g}_t;\bar{\mathbf{m}}_t;
\widetilde{\mathbf{h}}_i;d_{t,i};u_{t,i}],
\label{eq:candidate_features}
\end{equation}
where $\boldsymbol{\phi}_{t,i}\in\mathbb{R}^{2D+2M+4}$. The network produces
\begin{equation}
z_{t,i}=
\operatorname{MLP}_{\mathrm{sel}}
(\boldsymbol{\phi}_{t,i}),
\qquad
p_{t,i}=\sigma(z_{t,i}),
\label{eq:candidate_score}
\end{equation}
where $\sigma$ is the sigmoid function. The auxiliary score $p_{t,i}$ is used to compute statistics for STOP scoring. It is not the action probability.

\paragraph{STOP scoring.}
At each nonterminal step, $b_t>0$ and $\mathcal{R}_t\neq\varnothing$. The policy scores all remaining candidates and STOP. The STOP network summarizes the remaining candidates using
\begin{equation}
p_t^{\max}=
\max_{c_i\in\mathcal{R}_t}p_{t,i},
\qquad
\bar{p}_t=
\frac{1}{|\mathcal{R}_t|}
\sum_{c_i\in\mathcal{R}_t}p_{t,i},
\label{eq:score_statistics}
\end{equation}
and
\begin{equation}
u_t^{\max}=
\max_{c_i\in\mathcal{R}_t}u_{t,i},
\qquad
\bar{u}_t=
\frac{1}{|\mathcal{R}_t|}
\sum_{c_i\in\mathcal{R}_t}u_{t,i}.
\label{eq:complementarity_statistics}
\end{equation}

The STOP input combines these statistics with the selected content, temporal hits, and normalized budget:
\begin{equation}
\boldsymbol{\psi}_t=
[\mathbf{g}_t;\bar{\mathbf{m}}_t;b_t/K;
p_t^{\max};\bar{p}_t;u_t^{\max};\bar{u}_t].
\label{eq:stop_features}
\end{equation}
Here, $\boldsymbol{\psi}_t\in\mathbb{R}^{D+M+5}$. The STOP score is $z_t^{\mathrm{stop}}=\operatorname{MLP}_{\mathrm{stop}}(\boldsymbol{\psi}_t)$.

\paragraph{Action selection and state update.}
The available actions are $\mathcal{A}_t=\{\mathrm{STOP}\}\cup\mathcal{R}_t$. No candidate is hard-filtered by temporal or semantic relevance. All unselected candidates remain available while the selection budget permits; the policy alone chooses a candidate or STOP. We jointly normalize their scores:
\begin{equation}
\pi_{\Phi}(a\mid s_t)=
\frac{\exp(z_{t,a})}
{\exp(z_t^{\mathrm{stop}})
+\sum_{c_i\in\mathcal{R}_t}\exp(z_{t,i})},
\qquad a\in\mathcal{A}_t,
\label{eq:action_probability}
\end{equation}
where $z_{t,a}=z_t^{\mathrm{stop}}$ for STOP and
$z_{t,a}=z_{t,i}$ for candidate $c_i$.

During training, we sample actions from this distribution. At inference, we choose the action with the highest probability. Selecting $c_i$ gives $\mathcal{S}_{t+1}=\mathcal{S}_t\cup\{c_i\}$. We then update the remaining set, selected-content representation, temporal hits, and budget.

STOP leaves the set unchanged, $\mathcal{S}_{t+1}=\mathcal{S}_t$, and returns it. At the first step, this returns an empty set. Selection also ends when the budget is exhausted, the pool is empty, or $T_{\max}$ decisions have been made. After $T$ decisions, the final output is $\mathcal{S}=\mathcal{S}_T$.

\subsection{Two-Stage Training}\label{sec:train}

Supervised warm-start initializes candidate scoring before reinforcement learning (RL), mitigating the sparse reward for covering all required answer parts.

\paragraph{Stage 0: Supervised warm-start.}
We score all candidates at the initial state. At this state, the selected set is empty, so
$\mathbf{g}_0=\mathbf{0}$,
$\mathbf{m}_0=\mathbf{0}$, and $d_{0,i}=0$.
We train the initial-state sigmoid scores $p_{0,i}$ with weighted binary cross-entropy:
\begin{equation}
\mathcal{L}_{\mathrm{BCE}}
=
-\frac{1}{N}\sum_{i=1}^{N}
\left[
w_{+}y_i\log p_{0,i}
+
w_{-}(1-y_i)\log(1-p_{0,i})
\right],
\label{eq:warm_start_loss}
\end{equation}
where $y_i=1$ if $c_i$ is annotated as supporting at least one required answer part, and $y_i=0$ otherwise. We set $w_{+}$ to the ratio of negative to positive candidates in the training set and $w_{-}=1$.

This stage updates the encoder, feature fusion layer, self-attention layer, layer normalization, and candidate scoring network. It does not train the STOP network. We select the checkpoint with the highest mean average precision (MAP), computed by averaging per-question AP over answerable validation questions.

\paragraph{Stage 1: Reinforcement learning.}
We initialize the candidate representation and scoring modules from Stage 0. The STOP network is randomly initialized and trained in Stage 1. For each training question, the policy samples a trajectory
$(s_0,a_0,\ldots,s_{T-1},a_{T-1})$.
After termination, the selected set $\mathcal{S}$ receives the reward
\begin{equation}
R_{\mathrm{term}}(\mathcal{S})
=
\lambda R_{\mathrm{qa}}(\mathcal{S})
+\eta R_{\mathrm{time}}(\mathcal{S})
+\xi R_{\mathrm{cov}}(\mathcal{S})
-\gamma\operatorname{Cost}(\mathcal{S}),
\label{eq:terminal_reward}
\end{equation}
where $\lambda$, $\eta$, $\xi$, and $\gamma$ are non-negative weights.

\paragraph{Evidence sufficiency.}
An answerable question has $L$ required answer parts: one per query interval for Singleton questions, and one per required answer entity for Set questions. Let $\mathcal{G}_g\subseteq\mathcal{C}$ contain alternative candidates supporting part $g$; a candidate may support multiple parts.

The selected set is sufficient if it contains at least one candidate from every group:
\begin{equation}
R_{\mathrm{qa}}(\mathcal{S})
=
\begin{cases}
\displaystyle
\prod_{g=1}^{L}
\mathbb{I}[\mathcal{S}\cap\mathcal{G}_g\neq\varnothing],
& \text{if answerable},\\
\mathbb{I}[\mathcal{S}=\varnothing],
& \text{if unanswerable}.
\end{cases}
\label{eq:sufficiency_reward}
\end{equation}
For unanswerable questions, only an empty output is correct. This reward uses evidence annotations, not generated answers.

\paragraph{Temporal matching.}
We encourage selected candidates to match the query time. The reward is the mean overall temporal hit:
\begin{equation}
R_{\mathrm{time}}(\mathcal{S})
=
\begin{cases}
\displaystyle
\frac{1}{|\mathcal{S}|}
\sum_{c_i\in\mathcal{S}}\tau_i^{\max},
& \text{if } \mathcal{S}\neq\varnothing,\\
0, & \text{otherwise}.
\end{cases}
\label{eq:temporal_reward}
\end{equation}
This measures temporal matching, not factual consistency between candidates.

\paragraph{Query-interval coverage.}
For answerable questions, we encourage the selected evidence to hit the query intervals. Each interval is counted once:
\begin{equation}
R_{\mathrm{cov}}(\mathcal{S})
=
\frac{1}{|\mathcal{T}_q|}
\sum_{k=1}^{|\mathcal{T}_q|}
\mathbb{I}
\left[
\exists c_i\in\mathcal{S}:\tau_{i,k}=1
\right].
\label{eq:coverage_reward}
\end{equation}
This counts actual query intervals with any overlap, excluding padding; it does not measure full temporal coverage or answer sufficiency.

For answerable questions whose gold evidence is entirely untimed, we set $R_{\mathrm{cov}}(\mathcal{S})=0$. For unanswerable questions, we instead set
$R_{\mathrm{cov}}(\mathcal{S})
=\mathbb{I}[\mathcal{S}=\varnothing]$.
Thus, both $R_{\mathrm{qa}}$ and $R_{\mathrm{cov}}$ reward correct abstention on these questions.

\paragraph{Selection cost.}
We use $\operatorname{Cost}(\mathcal{S})=|\mathcal{S}|/K$ to penalize the number of selected items.

\paragraph{Policy optimization.}
We use REINFORCE~\cite{williams1992reinforce} with entropy regularization and a batch-mean baseline, $\bar{R}=B^{-1}\sum_{b=1}^{B}R_{\mathrm{term}}(\mathcal{S}^{(b)})$. Here, $B$ is the batch size and $\mathcal{S}^{(b)}$ is the sampled output for question $b$. We sample one trajectory per question.

The loss for one trajectory is
\begin{equation}
\begin{aligned}
\mathcal{L}_{\mathrm{RL}}
={}&
-\bigl(R_{\mathrm{term}}(\mathcal{S})-\bar{R}\bigr)
\sum_{t=0}^{T-1}\log\pi_{\Phi}(a_t\mid s_t)\\
&-\beta\sum_{t=0}^{T-1}
H\bigl(\pi_{\Phi}(\cdot\mid s_t)\bigr),
\end{aligned}
\label{eq:rl_loss}
\end{equation}
where $\Phi$ denotes all trainable model parameters, $\beta$ is the entropy weight, and $H(\pi)=-\sum_{a\in\mathcal{A}_t}\pi(a\mid s_t)\log\pi(a\mid s_t)$ is policy entropy.

Entropy regularization encourages exploration. Rewards and the baseline are detached; the batch-averaged loss updates all model parameters.

We select the Stage 1 checkpoint with the highest mean validation reward. Validation and inference use greedy actions.

\section{Experiments}
\label{sec:exp}

\subsection{Dataset Construction}
\label{sec:data}

Each question has $N=20$ candidate passages. \textbf{TempPool-TimeQA} follows TimeQA's~\cite{chen2021timeqa} train/\allowbreak dev/\allowbreak test splits; questions from the same fact stay in one split. \textbf{TempPool-SituatedQA} uses TempRAGEval's SituatedQA subset~\cite{zhang2025mrag} for cross-dataset testing. Both training stages use only the TimeQA training split.
\textbf{Singleton} questions require one answer group per query interval. \textbf{Set} questions require multiple distinct answer entities within one interval. \textbf{Unanswerable} questions have no supporting evidence under TC in the pool; their correct output is empty. Table~\ref{tab:data} reports the final counts.

\paragraph{TempPool-TimeQA.}
Each source fact records a subject's relation-specific answer timeline and a Wikipedia page. We sample TC from the timeline and insert a temporal phrase into a relation template to form $q$. Removing the phrase gives MC. The timeline determines the required answers and question type; candidate passages come from Wikipedia pages.

An LLM extracts candidate intervals from the passage and MC only, without TC, answers, or gold labels. Endpoints must be grounded in the passage; otherwise, it is treated as untimed. Separately, a passage is labeled \emph{gold} if it matches an answer string, overlaps TC or is untimed, and passes LLM answer-support verification. We discard an answerable question if any required answer lacks gold support.

We retain all gold and add negatives: subject-related but temporally conflicting passages (\emph{sibling}), temporally overlapping passages without answer support (\emph{date-bait}), and other same-page, untimed, or off-page passages. Pools are shuffled; questions with fewer than 20 candidates are discarded. Each fact yields at most one unanswerable question. We discard it if any candidate supports an answer under TC, regardless of whether time information is present. Unanswerable pools contain only non-supporting passages. TimeQA test pools average 1.51 gold, 2.64 temporally conflicting, and 1.29 temporally overlapping non-gold items.

\paragraph{TempPool-SituatedQA.}
We retain the wording of 500 source questions with temporal relations. MC precedes the last temporal relation. TC uses day-level intervals: \emph{as of/by/until} include the date; \emph{before/after} exclude it; \emph{since/from} include the start date. Bounded ranges are closed; \emph{in/on/around/during} denote the specified day or year. Gold passages must contain an annotated evidence sentence and pass temporal and LLM support checks. Questions missing required gold support are discarded, leaving 413. We retain their gold passages and fill the remaining pool slots with passages sampled randomly from other questions. Neither test set participates in training.

\begin{table}[t]
\centering
\small
\caption{Number of questions per split and type after construction and filtering. Avg.\ gold: mean number of gold items across all pools, including unanswerable ones.}
\label{tab:data}
\begin{tabular}{@{}l*{3}{>{\centering\arraybackslash}p{1.2cm}}c@{}}
\toprule
& \multicolumn{3}{c}{\textbf{TempPool-TimeQA}} & \textbf{TempPool-SituatedQA} \\
\cmidrule(lr){2-4}\cmidrule(lr){5-5}
& Train & Dev & Test & Test \\
\midrule
Total        & 7587 & 1139 & 2992 & 413 \\
Singleton    & 2004 & 304  & 791  & 413 \\
Set          & 3850 & 580  & 1536 & 0 \\
Unanswerable & 1733 & 255  & 665  & 0 \\
Avg.\ gold   & 1.41 & 1.37 & 1.51 & 1.61 \\
\bottomrule
\end{tabular}
\end{table}

\subsection{Experimental Setup}
\label{sec:setup}

We evaluate evidence quality, redundancy, and empty-set abstention on the same candidate pools. Evaluation uses evidence annotations rather than generated-answer accuracy.\footnote{Our code is available at \url{https://github.com/alan-chll/TASTE.git}.}

\paragraph{Baselines.}
Baselines form two groups.
The first group comprises four ranking baselines that produce a total order over the same 20-item pool and have no STOP action:

\begin{itemize}
\item \textbf{Contriever}~\cite{izacard2022contriever}: dual-encoder similarity scores between full $q$ and each pooled passage.
\item \textbf{BGE}~\cite{chen2024bge}: BGE-reranker-v2-m3, a cross-encoder reranker with full $q$.
\item \textbf{Qwen3}~\cite{zhang2025qwen3}: Qwen3-Reranker-4B, an LLM-based reranker with full $q$.
\item \textbf{MRAG}~\cite{zhang2025mrag}: we apply its semantic--temporal scoring module to the fixed pools, rather than its full retrieval and summarization pipeline. Our implementation uses NV-Embed-v2~\cite{lee2025nvembed} with temporal phrases removed from $q$, multiplied by a symbolic spline score over in-sentence years. This is a strong time-aware ranking baseline without selected-set conditioning or learned STOP. Percentile thresholds are our abstention protocol, not part of the original method.
\end{itemize}

Two adaptive-count methods use official configurations and the specified pretrained models, without our explicit temporal features:

\begin{itemize}
\item \textbf{DynRAG}~\cite{sun2025dynamicrag}: DynamicRAG-8B as reranker; full $q$ and 20 candidates are fed with the official prompt/chat template. The model generates selected IDs, or \texttt{None} when no evidence is needed. Decoding temperature 0.4; max generation length 100.
\item \textbf{Adapt-$k$}~\cite{taguchi2025adaptivek}: Adaptive-$k$ with similarities from \texttt{bge-large-en-v1.5}~\cite{xiao2024cpack} between full $q$ and candidates, sorted in descending order; the cut is placed at the largest adjacent-score drop among the top 90\% candidates, and five extra items are kept (official setting).
\end{itemize}

\paragraph{Model configuration.}
The encoder $\operatorname{Enc}$ in Eq.~\eqref{eq:semantic_relevance} is initialized from all-MiniLM-L6-v2~\cite{reimers2019sbert,wang2020minilm} ($D=384$, max length 128, mean pooling) and fine-tuned during training.
$\operatorname{Attn}$ in Eq.~\eqref{eq:pool_attention} uses 4 heads, and $\operatorname{MLP}_{\mathrm{sel}}$ and $\operatorname{MLP}_{\mathrm{stop}}$ are three-layer feed-forward networks with hidden size 256.
We set the pool size $N=20$, the maximum evidence count $K=10$, the maximum number of decision steps $T_{\max}=11$, and the maximum number of query intervals $M=4$.

\paragraph{Training configuration.}
Reward coefficients $\lambda=1$, $\eta=0.2$, $\xi=0.2$, and $\gamma=0.3$ are tuned on the validation set, and the entropy coefficient is $\beta=0.01$.
We use AdamW~\cite{loshchilov2019adamw} with learning rates of $5\times10^{-6}$ for $\operatorname{Enc}$ and $2\times10^{-5}$ for the other parameters, weight decay 0.01, batch size 8, and gradient clipping at 1.0.
Stage~0 and Stage~1 train for up to 20 and 12 epochs, respectively, with early stopping after 4 and 3 epochs without validation improvement under the model selection criteria in Section~\ref{sec:train}.
All training runs on a single NVIDIA A800 80GB GPU.

\paragraph{Comparison protocols.}
For the main quality comparison, each ranking baseline returns as many items as TASTE selects for the same question and seed, including zero. This isolates selection quality at matched counts, conditional on TASTE's count decisions. DynRAG and Adapt-$k$ use their own output counts. For abstention, rankers instead use score-percentile thresholds computed over all candidate scores in the test set (Table~\ref{tab:empty}); their matched-count outputs are not used.

\paragraph{Reporting.}
TASTE uses training seeds 0, 1, and 2. Main results report mean$\pm$std; ablations report means. Rankers are not retrained; matched-count variation reflects TASTE's seed-dependent counts. Cases use the seed-0 TASTE model.

\subsection{Evaluation Metrics}
\label{sec:metrics}

\noindent\textbf{Sufficiency rate (Suff.)} is the proportion of answerable questions with sufficient evidence, i.e., $R_{\mathrm{qa}}(\mathcal{S})=1$.

\smallskip
\noindent
\textbf{Useful rate (Useful)} checks items in output order. An item counts once if its gold annotations cover one or more previously uncovered answer parts. We average each question's fraction of useful items over answerable questions, assigning zero to empty outputs.

\smallskip
\noindent\textbf{Sufficiency--Usefulness score ($F_{\mathrm{SU}}$)} is $2\cdot\mathrm{Suff.}\cdot\mathrm{Useful}/(\mathrm{Suff.}+\mathrm{Useful})$, or zero when both terms are zero. These three metrics are computed per seed before reporting means and standard deviations.

\smallskip
\noindent\textbf{Conflict rate (Conflict)} is the fraction of selected items with $\tau_i^{\max}=-1$, pooled across all questions, including unanswerable ones. Untimed items are not conflicts; empty outputs contribute no items.

\smallskip
\noindent\textbf{Sufficient-prefix cost (Items, Tokens)} counts items and cumulative tokens up to the first sufficient prefix. Rankers use full 20-item orders. Table~\ref{tab:cost} retains TASTE's sufficient questions and excludes DynRAG/Adapt-$k$ failures; success sets may differ. This diagnostic measures neither final output length nor inference latency.

\smallskip
\noindent\textbf{Abstention rate (Abst.)} is the fraction of unanswerable questions returning $\varnothing$; the fraction of answerable questions returning $\varnothing$ is the \textbf{false abstention rate}.

\begin{table}[t]
\centering
\caption{Evidence selection results (\%, mean$\pm$std over three seeds; best in bold; Conflict is lower-is-better). Ranking baselines are truncated to TASTE's evidence counts.}
\label{tab:main}
\small
\setlength{\tabcolsep}{2.0pt}
\begin{adjustbox}{max width=\textwidth}
\begin{tabular}{@{}lcccccccc@{}}
\toprule
& \multicolumn{4}{c}{\textbf{TempPool-TimeQA}} & \multicolumn{4}{c}{\textbf{TempPool-SituatedQA}} \\
\cmidrule(lr){2-5}\cmidrule(lr){6-9}
\textbf{Method} & Suff. & Useful & $F_{\mathrm{SU}}$ & Conflict & Suff. & Useful & $F_{\mathrm{SU}}$ & Conflict \\
\midrule
Contriever  & 45.1{\scriptsize$\pm$4.1} & 25.8{\scriptsize$\pm$0.2} & 32.8{\scriptsize$\pm$1.0} & 18.5{\scriptsize$\pm$0.6} & 76.8{\scriptsize$\pm$4.8} & 37.7{\scriptsize$\pm$1.6} & 50.5{\scriptsize$\pm$0.6} & 1.9{\scriptsize$\pm$0.1} \\
BGE         & 73.6{\scriptsize$\pm$3.3} & 41.7{\scriptsize$\pm$1.2} & 53.2{\scriptsize$\pm$0.3} & 15.2{\scriptsize$\pm$0.9} & 84.2{\scriptsize$\pm$3.8} & 43.9{\scriptsize$\pm$2.3} & 57.6{\scriptsize$\pm$1.3} & 2.0{\scriptsize$\pm$0.1} \\
Qwen3       & 80.7{\scriptsize$\pm$3.1} & 46.5{\scriptsize$\pm$1.7} & 59.0{\scriptsize$\pm$0.6} & 12.6{\scriptsize$\pm$1.1} & 90.6{\scriptsize$\pm$4.4} & 48.0{\scriptsize$\pm$1.8} & 62.7{\scriptsize$\pm$1.1} & 1.7{\scriptsize$\pm$0.4} \\
MRAG        & 75.6{\scriptsize$\pm$3.2} & 44.4{\scriptsize$\pm$1.6} & 55.9{\scriptsize$\pm$0.5} & 10.7{\scriptsize$\pm$1.2} & 88.8{\scriptsize$\pm$5.2} & 46.9{\scriptsize$\pm$2.0} & 61.3{\scriptsize$\pm$1.8} & 2.1{\scriptsize$\pm$0.1} \\
DynRAG      & 98.1{\scriptsize$\pm$0.4} & 17.5{\scriptsize$\pm$0.2} & 29.7{\scriptsize$\pm$0.3} & 18.4{\scriptsize$\pm$0.2} & 99.7{\scriptsize$\pm$0.3} & 15.3{\scriptsize$\pm$0.5} & 26.5{\scriptsize$\pm$0.8} & 2.1{\scriptsize$\pm$0.1} \\
Adapt-$k$   & \textbf{99.7}{\scriptsize$\pm$0.0} & 11.9{\scriptsize$\pm$0.0} & 21.3{\scriptsize$\pm$0.0} & 17.9{\scriptsize$\pm$0.0} & \textbf{100.0}{\scriptsize$\pm$0.0} & 9.5{\scriptsize$\pm$0.0} & 17.4{\scriptsize$\pm$0.0} & 1.8{\scriptsize$\pm$0.0} \\
TASTE        & 92.0{\scriptsize$\pm$2.3} & \textbf{55.8}{\scriptsize$\pm$2.4} & \textbf{69.4}{\scriptsize$\pm$1.2} & \textbf{0.6}{\scriptsize$\pm$0.9} & 93.4{\scriptsize$\pm$4.5} & \textbf{50.7}{\scriptsize$\pm$1.9} & \textbf{65.6}{\scriptsize$\pm$1.1} & \textbf{0.1}{\scriptsize$\pm$0.0} \\
\bottomrule
\end{tabular}
\end{adjustbox}
\end{table}

\begin{table}[t]
\begin{minipage}[t]{0.48\textwidth}
\centering
\caption{Sufficient-prefix cost (mean$\pm$std over three seeds; lower is better). Evaluated where TASTE suffices; DynRAG/Adapt-$k$ failures excluded.\strut}
\label{tab:cost}
\footnotesize
\setlength{\tabcolsep}{2.0pt}
\renewcommand{\arraystretch}{1.185}
\begin{adjustbox}{max width=\linewidth}
\begin{tabular}{@{}lcccc@{}}
\toprule
& \multicolumn{2}{@{}c@{}}{\fontsize{6}{7}\selectfont\textbf{TempPool-TimeQA}} & \multicolumn{2}{@{}c@{}}{\fontsize{6}{7}\selectfont\textbf{TempPool-SituatedQA}} \\
\cmidrule(lr){2-3}\cmidrule(lr){4-5}
\textbf{Method} & Items & Tokens & Items & Tokens \\
\midrule
Contriever & 4.2{\scriptsize$\pm$0.0} & 274.1{\scriptsize$\pm$1.1} & 1.7{\scriptsize$\pm$0.0} & 265.0{\scriptsize$\pm$0.5} \\
BGE        & 2.6{\scriptsize$\pm$0.0} & 201.0{\scriptsize$\pm$2.0} & 1.5{\scriptsize$\pm$0.0} & 271.7{\scriptsize$\pm$1.8} \\
Qwen3      & 2.4{\scriptsize$\pm$0.0} & 172.4{\scriptsize$\pm$1.8} & \textbf{1.2}{\scriptsize$\pm$0.0} & \textbf{195.7}{\scriptsize$\pm$1.8} \\
MRAG       & 2.7{\scriptsize$\pm$0.0} & 210.3{\scriptsize$\pm$2.5} & 1.3{\scriptsize$\pm$0.0} & 226.2{\scriptsize$\pm$1.8} \\
DynRAG     & 2.4{\scriptsize$\pm$0.1} & 196.7{\scriptsize$\pm$4.5} & 1.3{\scriptsize$\pm$0.0} & 226.1{\scriptsize$\pm$5.8} \\
Adapt-$k$  & 3.0{\scriptsize$\pm$0.0} & 215.6{\scriptsize$\pm$2.1} & 1.3{\scriptsize$\pm$0.0} & 224.7{\scriptsize$\pm$5.3} \\
TASTE       & \textbf{2.0}{\scriptsize$\pm$0.1} & \textbf{160.8}{\scriptsize$\pm$5.4} & \textbf{1.2}{\scriptsize$\pm$0.1} & 204.6{\scriptsize$\pm$15.5} \\
\bottomrule
\end{tabular}
\end{adjustbox}
\end{minipage}\hfill
\begin{minipage}[t]{0.48\textwidth}
\centering
\caption{Abstention on unanswerable TimeQA questions (\%, mean$\pm$std). P$xx$: ranking baselines keep only candidates scoring above the $xx$-th percentile of all candidate scores; others use their own outputs.\strut}
\label{tab:empty}
\footnotesize
\setlength{\tabcolsep}{2.0pt}
\begin{adjustbox}{max width=\linewidth}
\begin{tabular}{@{}lcccc@{}}
\toprule
& Contriever & BGE & Qwen3 & MRAG \\
\midrule
P60 & 0.0{\scriptsize$\pm$0.0} & 0.0{\scriptsize$\pm$0.0} & 0.0{\scriptsize$\pm$0.0} & 0.0{\scriptsize$\pm$0.0} \\
P70 & 1.7{\scriptsize$\pm$0.0} & 2.3{\scriptsize$\pm$0.0} & 0.3{\scriptsize$\pm$0.0} & 3.3{\scriptsize$\pm$0.0} \\
P80 & 11.1{\scriptsize$\pm$0.0} & 11.0{\scriptsize$\pm$0.0} & 7.1{\scriptsize$\pm$0.0} & 19.8{\scriptsize$\pm$0.0} \\
P90 & 33.8{\scriptsize$\pm$0.0} & 35.8{\scriptsize$\pm$0.0} & 34.3{\scriptsize$\pm$0.0} & 57.3{\scriptsize$\pm$0.0} \\
P95 & 58.3{\scriptsize$\pm$0.0} & 59.4{\scriptsize$\pm$0.0} & 61.1{\scriptsize$\pm$0.0} & 70.4{\scriptsize$\pm$0.0} \\
\bottomrule
\end{tabular}
\end{adjustbox}\\[0.35em]
\begin{adjustbox}{max width=\linewidth}
\begin{tabular}{@{}lccc@{}}
& DynRAG & Adapt-$k$ & TASTE \\
\midrule
Original & 0.1{\scriptsize$\pm$0.1} & 0.0{\scriptsize$\pm$0.0} & \textbf{80.9}{\scriptsize$\pm$3.1} \\
\bottomrule
\end{tabular}
\end{adjustbox}
\end{minipage}
\end{table}

\section{Results and Analysis}
\label{sec:results}

\subsection{Baseline Comparison}
\label{sec:main}

Tables~\ref{tab:main}--\ref{tab:empty} compare TASTE with ranking and adaptive-count baselines.

Table~\ref{tab:main} shows that TASTE improves sufficiency, useful rate, and $F_{\mathrm{SU}}$ over all ranking baselines at matched evidence counts. The gains on the singleton-only TempPool-SituatedQA set also indicate cross-dataset transfer. DynRAG and Adapt-$k$ achieve higher sufficiency, but lower useful rates and $F_{\mathrm{SU}}$. Their lower useful rates indicate a larger proportion of redundant or non-supporting items. TASTE's conflict rates are 0.6\% and 0.1\%, respectively, below all baselines.

\paragraph{Time-aware ranking.}
MRAG combines strong semantic representations with explicit temporal scoring. On TempPool-TimeQA, its sufficiency, useful rate, and $F_{\mathrm{SU}}$ are 75.6\%, 44.4\%, and 55.9\%, versus TASTE's 92.0\%, 55.8\%, and 69.4\%, at matched per-question counts for each seed. Thus, TASTE also outperforms time-aware ranking. This comparison uses MRAG's scoring module, not its full pipeline, and does not isolate temporal features, sequential selection, or stopping.

Table~\ref{tab:cost} reports conditional sufficient-prefix costs. TASTE has the lowest item and token counts on TimeQA. On SituatedQA, Qwen3 and TASTE both average 1.2 items, but Qwen3 uses fewer tokens (195.7 versus 204.6). These costs exclude items after sufficiency and do not establish an inference-time advantage.

Table~\ref{tab:empty} covers 665 unanswerable TimeQA test questions. TASTE abstains on 80.9\%, exceeding all rankers even at the strictest tested threshold, P95 (at most 70.4\%). DynRAG and Adapt-$k$ rarely abstain. These rates are not matched for false abstention. TASTE's false abstention rate, 2.4\%$\pm$0.9, pools answerable questions across both datasets, rather than TimeQA alone.

\subsection{Ablation Study}
\label{sec:ablation}
\begin{figure}[t]
    \centering
    \includegraphics[width=0.95\linewidth]{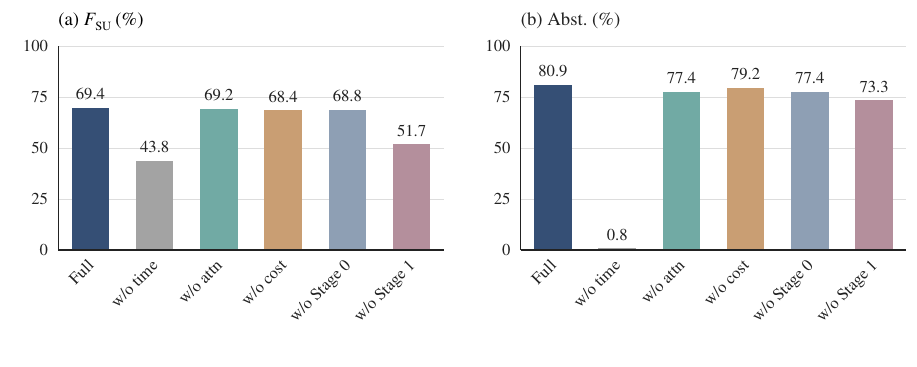}
    \caption{Ablations on TempPool-TimeQA (three-seed means; higher is better): left, $F_{\mathrm{SU}}$ on answerable questions; right, abstention on unanswerable questions.}
    \label{fig:ablation}
\end{figure}

Figure~\ref{fig:ablation} reports $F_{\mathrm{SU}}$ and Abst. for ablations on the TimeQA test pools.

\paragraph{(1) Temporal and attention ablations.}
For \textbf{w/o time}, we disable all explicit temporal information: candidate hits $\tau$, the missing-time flag $\mu_i$, selected-set hits $\mathbf{m}_t$, and the time-dependent complementarity feature $u_{t,i}$. We also remove $R_{\mathrm{time}}$ and $R_{\mathrm{cov}}$ from the reward. For \textbf{w/o attn}, we remove self-attention but keep feature fusion. Both variants are retrained through Stage~0 and Stage~1.

\paragraph{(2) Reward ablation.}
For \textbf{w/o cost}, we remove the cost penalty from the terminal reward but keep STOP, then retrain Stage~1.

\paragraph{(3) Training ablations.}
For \textbf{w/o Stage~0}, we skip supervised warm-start and randomly initialize the selection and stopping heads for Stage~1 training.
For \textbf{Stage~0 only (w/o Stage~1)}, we keep the supervised warm-start ranker and omit reinforcement learning, so there is no learned STOP.
For $F_{\mathrm{SU}}$, top-$k$ uses $k$ determined separately per seed from Full's average evidence count, unlike per-question matching in Table~\ref{tab:main}. For Abst., we choose a Stage~0 score-percentile threshold matching Full's false-abstention rate on answerable test questions, then apply it to unanswerable test questions. This is a retrospective matched-rate comparison, not independently calibrated test performance.

Removing the explicit temporal mechanisms, including time-related features and rewards, reduces $F_{\mathrm{SU}}$ from 69.4\% to 43.8\% and Abst. from 80.9\% to 0.8\%. Removing self-attention, the cost penalty, or Stage~0 lowers $F_{\mathrm{SU}}$ by at most 1.0 points and Abst. by at most 3.5 points. The temporal mechanisms jointly have the largest measured effect. The smaller changes for the other components provide limited evidence of their individual contributions under this setup.

Stage~0 only obtains $F_{\mathrm{SU}}=51.7\%$ and Abst.\ $=73.3\%$, compared with 69.4\% and 80.9\% for Full. Stage~1 improves $F_{\mathrm{SU}}$ by 17.7 points and Abst.\ by 7.6 points under these protocols. These gains reflect Stage~1 as a whole; they do not isolate selection from STOP learning.

\subsection{Case Study}
\label{sec:case}

\begin{table}[H]
\centering
\small
\caption{Rankings and TASTE actions for three cases. Candidate IDs $c_i$ are local to each pool; $\varnothing$ means no supporting evidence. Rankings run top to bottom; ellipses omit ranks. Arrows show successive actions.}
\label{tab:case}
\setlength{\tabcolsep}{2pt}
\begin{tabular}{@{}>{\raggedright\arraybackslash}p{0.34\textwidth}>{\raggedright\arraybackslash}p{0.17\textwidth}>{\raggedright\arraybackslash}p{0.14\textwidth}>{\raggedright\arraybackslash}p{0.13\textwidth}>{\raggedright\arraybackslash}p{0.15\textwidth}@{}}
\toprule
\textbf{Question} & \textbf{Required support} & \textbf{Qwen3}\newline Ranking & \textbf{Stage~0}\newline Ranking & \textbf{TASTE}\newline Actions \\
\midrule
Q1: What was Rush Holt Jr.'s occupation on 26 Nov 1983?
& $c_{11}$
& $c_{16}$\newline $\ldots$\newline $c_{11}$
& $c_{11}$\newline $c_{6}$\newline $c_{13}$
& $c_{11}\rightarrow$\newline $\mathrm{STOP}$ \\
\midrule
Q2: Who did Katherine O'Brien work for before 2006?
& $c_{3}$, $c_{14}$, $c_{9}$
& $c_{14}$\newline $c_{4}$\newline $c_{9}$\newline $\ldots$\newline $c_{3}$ (rank 8)
& $c_{14}$\newline $c_{3}$\newline $c_{13}$\newline $c_{9}$
& $c_{3}\rightarrow c_{14}$\newline $\rightarrow c_{9}$\newline $\rightarrow\mathrm{STOP}$ \\
\midrule
Q3: What was the nationality of Alexander Imich on 22 Jun 1922?
& $\varnothing$
& $c_{15}$
& $c_{15}$\newline $c_{18}$\newline $c_{9}$
& $\mathrm{STOP}$ \\
\bottomrule
\end{tabular}
\end{table}

Table~\ref{tab:case} compares Qwen3, Stage~0, and full TASTE on seed~0.

\noindent\textbf{Temporal validity and stopping (Q1, singleton).}
Qwen3 ranks $c_{16}$ (U.S.\ Representative, 1999--2015) first, although the question concerns 1983. Stage~0 ranks $c_{11}$ (faculty at Swarthmore, 1980--1988) first. TASTE selects $c_{11}$ and stops once this support is sufficient.

\noindent\textbf{Complementary selection (Q2, set).}
Each gold item provides required support that the other two miss. Qwen3 places $c_{3}$ eighth. Stage~0 ranks the non-gold $c_{13}$ above $c_{9}$, which adds support after Jul~1998. TASTE selects $c_{3}$ and $c_{14}$, adds $c_{9}$ to complete support, and stops.

\noindent\textbf{Abstention (Q3, unanswerable).}
Candidate $c_{15}$ describes Imich as Polish-American, but his birth year (1903) does not establish nationality on the queried date. Stage~0 still provides a ranking; TASTE takes STOP before selecting any item and returns an empty set.

\section{Conclusion}
\label{sec:conclusion}

TASTE jointly learns evidence selection and stopping for time-sensitive QA. Its policy uses semantic and temporal features and the selected evidence to choose the next item or STOP. On two fixed-pool datasets, TASTE improves sufficiency and usefulness over ranking baselines at matched evidence counts and reduces temporal conflicts. It returns empty sets for 80.9\% of unanswerable TempPool-TimeQA test questions. Future work will evaluate generated-answer quality and candidate pools from real retrieval systems.

\bibliographystyle{splncs04}
\bibliography{references}

\end{document}